\pdfoutput=1

\documentclass[11pt]{article}

\usepackage[final]{emnlp2021}

\usepackage{times}
\usepackage{latexsym}

\usepackage[T1]{fontenc}

\usepackage[utf8]{inputenc}

\usepackage{booktabs}
\usepackage{microtype}
\usepackage{graphicx}
\usepackage{xcolor}
\usepackage{framed}
\usepackage{caption}
\usepackage{xspace}
\usepackage{subcaption}
\usepackage{multirow}
\usepackage{tablefootnote}
\usepackage{pythonhighlight}
\usepackage{enumitem}
\usepackage{footnote}

\newcommand{\name}{\textsc{\textsf{TuringBench}}}

\usepackage{microtype}
\usepackage{adjustbox}

\title{{\name}: A Benchmark Environment for Turing Test \\in the Age of Neural Text Generation}

\author{Adaku Uchendu\hspace{0.2in}
        Zeyu Ma$^{\dagger}$\hspace{0.2in}
        Thai Le \hspace{0.2in}
        Rui Zhang \hspace{0.2in}
        Dongwon Lee \vspace{0.1in} \\
        The Pennsylvania State University, University Park, PA, USA \\
        \texttt{\{azu5030, thaile, rmz5227, dongwon\}@psu.edu} \vspace{0.1in} \\
        Carnegie Mellon University, Pittsburgh, PA, USA$^{\dagger}$ \\
        \texttt{mazeyuwhu@gmail.com}$^{\dagger}$
}

\begin{document}

\maketitle
\begin{abstract}

Recent progress in generative language models has enabled machines to generate astonishingly realistic texts. While there are many legitimate applications of such models, there is also a rising need to distinguish machine-generated texts from human-written ones (e.g., fake news detection). However, to our best knowledge, there is currently no benchmark environment with datasets and tasks to systematically study the so-called 
``Turing Test'' problem for neural text generation methods.
In this work, we present the {\name} 
benchmark environment, which 
is comprised of (1) a dataset with 200K human- or machine-generated samples across 20 labels 
\{Human, 
GPT-1, GPT-2\_small, GPT-2\_medium, GPT-2\_large, 
GPT-2\_xl, GPT-2\_PyTorch, GPT-3, GROVER\_base, GROVER\_large,
GROVER\_mega, CTRL, XLM, XLNET\_base, XLNET\_large, FAIR\_wmt19, 
FAIR\_wmt20, TRANSFORMER\_XL, PPLM\_distil, PPLM\_gpt2\}, (2) two 
benchmark tasks--i.e., \textit{Turing Test} (TT) and \textit{Authorship Attribution} (AA), and (3) a website with leaderboards. 
Our preliminary experimental results using  {\name}  show that
FAIR\_wmt20 and  GPT-3 are the current winners, among all language models tested, in generating the most human-like indistinguishable texts with the lowest F1 score by five state-of-the-art TT detection models.
The {\name} is available at:
\textcolor{blue}{\url{https://turingbench.ist.psu.edu/}}

\end{abstract}

\section{Introduction}

\begin{figure}[!htb]
  \centering
  \fbox{%
    \begin{tabular}{p{0.44\textwidth}}
      \noindent
\small
\textcolor{orange}{\textbf{PROMPT}:} \textit{Some investigators struggle to reach Comey}
\\\hline
\small
\textcolor{blue}{\textbf{HUMAN}:} Some congressional investigators want to bring former FBI Director James Comey to the Hill to tell them about his meetings with President Donald Trump but they can’t find him. 
\\
\small
\textcolor{cyan}{\textbf{GPT-2}:} Some investigators struggle to reach Comey and his team because the FBI's headquarters is in Northern Virginia. But the FBI has a lot of other people who can help.
\\
\small
\textcolor{purple}{\textbf{FAIR}:} We have a duty to be vigilant. James Comey's testimony today should make him think twice about it. The White House said Trump had spoken with Comey during the hour-long meeting and that all the senior U.S. officials were given a "heads up" on the situation.
\\
\small
\textcolor{magenta}{\textbf{GPT-3}:} ``Like Louis XVI, he doesn’t see the storm growing in the distance,'' says the Democratic operative. The lack of specifics, even from surrogates on Trump’s behalf, forces well-known Democrats to point out the obvious.
    \end{tabular}}
  \caption{Snippets of \textcolor{orange}{\textbf{PROMPT}} (a news article title), \textcolor{blue}{\textbf{HUMAN}} vs. three examples of machine-generated texts (\textcolor{cyan}{\textbf{GPT-2}}, \textcolor{purple}{\textbf{FAIR}} and \textcolor{magenta}{\textbf{GPT-3}}).}
  \label{fig:example}
\end{figure}

\begin{figure}
    \centering
    \includegraphics[width=1.05\linewidth]{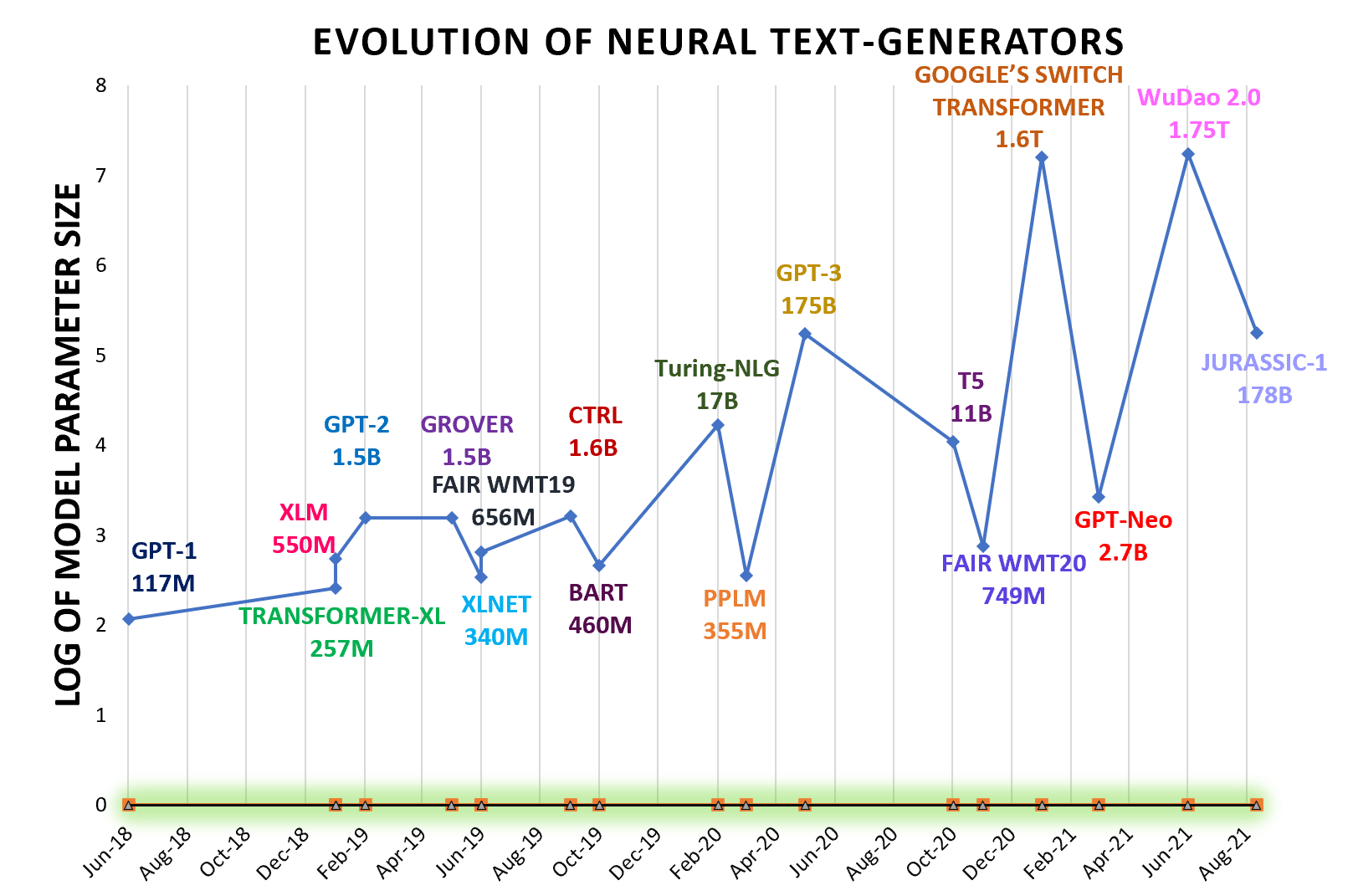}
    \caption{Evolution of neural text generators ($Y$-axis depicts model parameters in millions in log plot).} 
    \label{fig:evolve}
\end{figure}

Recently, the field of Natural Language Generation (NLG) has seen a massive improvement.
While the field of NLG has existed for some time since even before the onset of the first chatbot ELIZA \cite{weizenbaum1966eliza}, the recent neural architecture Transformers \cite{vaswani2017attention} has led to speedy improvement in the generation of long coherent texts. 
GPT-1 \cite{radford2018improving} created by OpenAI is the first installment of these new waves of text-generators. 
In fact, GPT-1 was built with 117 million parameters, however,
in less than 3 years, 
Google's Switch Transformer \citep{fedus2021switch} was the largest language model with 1.6 trillion parameters as of January-June 2021. Currently, the largest language model is 
Beijing Academy of Artificial Intelligence's (BAAI) WuDao 2.0 with 1.75 trillion parameters.
Even more alarming, since the birth of GPT-1, the field of NLG has grown exponentially such that
Hugging Face's model repo houses more than 9K English and non-English
language models (of which over 2K are text-generators). See Figure \ref{fig:evolve} for evolution 
of neural text-generators. 
Naturally, these newer language models are
able to generate texts that can be easily misconstrued as human-written.
Thus, due to the superior quality of recent generated texts 
and how easily such text-generators can be used, 
the potential for misuse is great. This misuse includes 
but is not limited to the spread of \textit{misinformation} \cite{zellers2019defending} and 
\textit{political propaganda} \cite{varol2017online}.
Therefore, it is 
urgent that we tackle ways to automatically distinguish machine-generated texts from human-written ones accurately.

To build accurate detectors of machine-generated texts, sufficient data is required but lacking. 
Therefore, we create a benchmark environment, {\name}, to combat the obvious security issue language models could pose. 
Just in line with benchmark environments such as SQuAD 
\cite{rajpurkar-etal-2016-squad} and GLUE \cite{wang-etal-2018-glue} that tremendously facilitate the 
progress of Natural Language Understanding, we build the first benchmark for Authorship Attribution 
in the form of the Turing Test by including humans and neural language models.

The {\name} Environment comprises benchmark datasets, benchmark tasks, 
and a website to host leaderboards.
This benchmark dataset is created by collecting 10K news articles 
(mostly in politics) written by journalists in 
media outlets such as CNN, Washington Post, etc.
Using the \textcolor{red}{Title} of each 
article, we \textcolor{orange}{Prompt} 19 selected neural text-generators to generate an article similar 
to the human-written one. This creates 200K articles with 20 labels (or authors). 
Next, we have two benchmark tasks 
- \textit{Turing Test}  and \textit{Authorship Attribution}. The \textit{Turing Test} task is modeled after the \textit{Turing Test} concept~\cite{turing2009computing}, where 
if a machine shows intelligent behavior or characteristics usually attributed to a human, then 
the machine has passed the test. In this scenario, the goal is to cause the machine to fail the \textit{Turing Test}. 
Thus, we define this benchmark task as a binary classification problem with 
\textit{human} and \textit{machine} labels. 
Given 19 neural text-generators, there are 19 \textit{Turing Test} subtasks with 19 human-machine pairs.

Furthermore, we understand that due to the ubiquitous nature of these 
neural language models, 
distinguishing machine-generated texts from human-written ones is no longer sufficient. 
It is now also important we inquire as to which particular neural text-generator authored a piece of text.
To this end, the \textit{Authorship Attribution} task aims to assign authorship to one of the many text-generators.
We study 20 authors for this task, however, as we have observed,
this can easily become 2K authors very soon which will grossly exacerbate the difficulty of this task.
Finally, to host all these tasks and datasets, 
we build a {\name} website with leaderboards for each benchmark task 
and call for participation in tackling this very relevant and non-trivial problem.

Lastly, we compare State-of-the-art (SOTA) and baseline \textit{Turing Test} and \textit{Authorship Attribution} models.
From the experimental results, we observe that we need more complex models to accurately distinguish 
machine-generated texts from human-written ones, including text-generators that are yet to be created.

\section{Related Work}

\paragraph{Neural Text Generation}
Recent advances in neural network-based language modeling have demonstrated promising results in text generation~\cite{garbacea2020neural}.
Current state-of-the-art neural text generation models can produce texts approaching the quality of human-written ones, especially in terms of grammar, fluency, coherency, and usage of real world
knowledge~\cite{radford2018improving,radford2019language,keskar2019ctrl,zellers2019defending,deng2020residual,brown2020language}.
The progress in neural text generation has facilitated a wide range of applications: dialog response generation~\cite{zhang2019dialogpt}, storytelling~\cite{fan2018hierarchical,see2019massively}, table-to-text generation~\cite{lebret-etal-2016-neural}, code comment generation~\cite{alon2018code2seq}, medical report generation~\cite{liu2019clinically}.

However, as these language models can generate text indistinguishable from human-written text, they can also be misused by adversaries to generate fake news~\cite{shu2017fake,wang2017liar,zellers2019defending,mosallanezhad2020topic,shu2020fact}, fake produce reviews~\cite{fornaciari-poesio-2014-identifying,adelani2020generating}, spam emails~\cite{das2019automated}.

\paragraph{Automatic Detection of Generated Text}
Given the potential malicious applications of text generation~\cite{solaiman2019release}, it is thus vital to build detectors to distinguish text generated by machines from humans~\cite{gehrmann-etal-2019-gltr,bakhtin2019real,jawahar2020automatic,varshney2020limits,ccano2020human}.
Most current work focus on fake news detection~\cite{rashkin-etal-2017-truth,zhou2019fake,bhat2020effectively,zhong-etal-2020-neural,schuster2020limitations,ippolito-etal-2020-automatic}.
Despite this progress, it remains a challenging task to build generalizable, interpretable, and robust detectors~\cite{jawahar2020automatic}.

\paragraph{Authorship Attribution}
Authorship Attribution (AA) aims to decide the author of a given text from a set of candidates~\cite{houvardas2006n,stamatatos2009survey,zhang2014authorship}.
AA has a broad range of applications including author profiling~\cite{lopez2020early}, computer forensics~\cite{lambers2009forensic}, and plagiarism detection~\cite{stamatatos2009intrinsic}.
Previous work on AA has explored and combined various features and representations at different levels including n-grams~\cite{escalante2011local,sapkota2015not,sapkota2016domain}, POS-tags~\cite{ferracane2017leveraging,halvani2020improved} psycholinguistics features~\cite{li2014towards,uchendu2019characterizing}, while recent approaches also build deep neural network based classifiers such as feed-forward NNLMs~\cite{ge2016authorship}, CNNs~\cite{hitschler2017authorship,shrestha2017convolutional}, LSTMs~\cite{jafariakinabad2019style,jafariakinabad2020self}, and BERT-based models~\cite{uchendu-etal-2020-authorship}.

However, previous AA work largely focuses on authorship attribution among humans, while only a few papers~\cite{manjavacas2017assessing,uchendu-etal-2020-authorship,munir2021through} study neural generated text.
Our work aims to provide the first benchmark for Authorship Attribution in the form of the Turing Test by including humans and neural language models.

\begin{table*}[th!]
\footnotesize
\resizebox{\textwidth}{!}{
\begin{tabular}{p{0.25\linewidth} p{0.75\linewidth}}
\toprule
\textbf{Text Generator} & \multicolumn{1}{c}{\textbf{Description}} \\
\midrule
Human  & We collected news titles (mostly Politics) and contents from CNN, Washington Post, and Kaggle. 
The Kaggle datasets had news articles from 2014--2020, and 2019--2020 for the CNN and Washington Post news articles. 
Next, we removed articles that did not have the desired word length (i.e., 200--500). This resulted in 130K articles, but only 10K was used for the article generations.
See data generation process in Figure \ref{fig:datagen}.
\\ \midrule
GPT-1    &  
Texts are generated with the Hugging Face github repo \cite{wolf2019huggingface}.\\
GPT-2   & We use 5 GPT-2 pre-trained models - \textbf{PyTorch} model, \textbf{small} (124 million parameters), \textbf{medium} (355 million parameters), \textbf{large} (774  million parameters), and \textbf{x-large} (1558 million  parameters) to generate texts.  \\
GPT-3 & Texts are generated with the OpenAI GPT-3 API using the \textit{davinci} engine. \\
GROVER  & We use code from repo to generate from Grover's 3 pre-trained models: \textbf{GROVER-base}, \textbf{GROVER-large}, \textbf{GROVER-mega}. \\
CTRL & Conditional Transformer Language Model For Controllable Generation uses control codes to guide generation. We use \textit{News} control code to generate long articles.\\
XLM & We generated texts using Hugging Face repo \cite{wolf2019huggingface}. \\
XLNET  & We generated texts with: 2 XLNET pre-trained models: \textbf{XLNET-base}, and \textbf{XLNET-large} using Hugging Face.\\
FAIR\_wmt & We use two Facebook's FAIR English models - \textbf{wmt19} \citep{ng2019facebook}  and  \textbf{wmt20} \cite{chen2020facebook} to generate texts with FAIRSEQ sequence modeling toolkit.\\
TRANSFORMER\_XL & We generated texts with this language model's setup on Hugging Face \cite{wolf2019huggingface}.\\
PPLM & PPLM fuses GPT-2's pre-trained model with bag of words to generate more specific texts. We used the \textit{Politics} bag of words model to generate texts. Next, we fused PPLM with two pre-trained models (i.e., distilGPT-2, and GPT-2) and generated texts with them, forming: \textbf{PPLM\_distil}, \textbf{PPLM\_gpt2}. These models are gotten from the Hugging Face model repository. \\
\bottomrule
\end{tabular}
}
\caption{Description of the text generators in the \name~dataset.}
\label{tab:desc}
\end{table*}
\section{The {\name} Environment}

\begin{figure}[t!]
    \centering
    \includegraphics[width=1.05\linewidth]{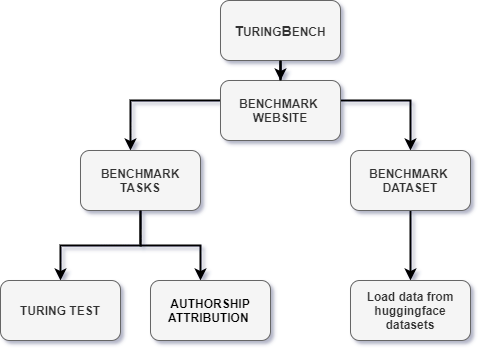}
    \caption{The \name~Environment.}
    \label{fig:turing}
\end{figure}

\begin{figure*}[t!]
    \centering
    \includegraphics[width=1\linewidth]{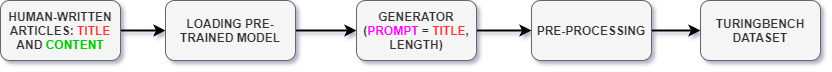}
    \caption{The \name~Data Collection, Generation, and Building process.}
    \label{fig:datagen}
\end{figure*}

Figure~\ref{fig:turing} overviews the framework of the {\name} Environment.

\subsection{Chosen Language Models} \label{datagen}
We generated texts using 10 language model architectures - 
\textit{GPT-1}~\citep{radford2018improving}, 
\textit{GPT-2}~\citep{radford2019language}, 
\textit{GPT-3}~\citep{brown2020language},
\textit{GROVER}~\citep{zellers2019defending}, 
\textit{CTRL}~\citep{keskar2019ctrl}, 
\textit{XLM}~\citep{lample2019cross}, 
\textit{XLNET}~\citep{yang2019xlnet}, 
\textit{FAIR}~\citep{ng2019facebook, chen2020facebook},
\textit{TRANSFORMER-XL}~\citep{dai2019transformer}, and
\textit{PPLM}~\citep{dathathri2019plug}.
In addition, some of these language models have multiple pre-trained models and thus, 
we were able to generate texts with 19 neural machine text-generators. 
We choose these 10 language model architectures because they are currently considered as the SOTA text-generators, many of the text-generators on Hugging Face's model repo are variants of these
language models,
and both their pre-trained models and codes were publicly available. 

To generate texts, all 19 neural generators require a short prompt and a specified number of words to generate texts. 
Table~\ref{tab:desc} (and Appendix) describes each language model in detail.
Figure~\ref{fig:datagen} illustrates the data creation process.
Table~\ref{tab:summary} summarizes the stats of dataset and the model sizes.



\begin{table*}[t!]
\centering
\footnotesize
\resizebox{14cm}{!}{
\begin{tabular}{lccc}
\toprule
\textbf{Text Generator}  & \textbf{\# of words (AVG $\pm$ Std. Dev.)}  & \textbf{\# of sentences (AVG $\pm$ Std. Dev.)} &  \textbf{ Model Parameter Size}\\
\midrule
\textbf{Human}  &  232.7 $\pm$ 42.0 & 15.0 $\pm$ 6.6 & N/A \\
\textbf{GPT-1} & 316.7 $\pm$ 12.9 & 10.5 $\pm$ 3.7 & 117M  \\
\textbf{GPT-2\_small} &  118.6 $\pm$ 61.0 &  4.0 $\pm$ 3.8 & 124M  \\
\textbf{GPT-2\_medium} &  120.9 $\pm$  66.0 & 4.2 $\pm$ 3.7 & 355M\\
\textbf{GPT-2\_large} & 119.7 $\pm$ 62.1 & 4.1 $\pm$  3.8 & 774M \\
\textbf{GPT-2\_xl} & 117.8 $\pm$ 63.3 & 4.1 $\pm$ 3.8 & 1.5B\\
\textbf{GPT-2\_PyTorch} & 178.9 $\pm$ 55.4 & 7.03 $\pm$ 4.8 & 344M \\
\textbf{GPT-3} &  129.5 $\pm$ 54.9 &  5.0 $\pm$ 3.7 & 175B\\
\textbf{GROVER\_base} & 299.2  $\pm$ 108.6 & 9.4 $\pm$ 6.9 & 124M\\
\textbf{GROVER\_large}  & 286.3 $\pm$ 101.3 &  8.7 $\pm$ 5.9 & 355M \\
\textbf{GROVER\_mega}  & 278.9 $\pm$ 97.6 & 9.2 $\pm$ 6.1 & 1.5B\\
\textbf{CTRL} & 398.1 $\pm$  64.8 & 20.0 $\pm$ 10.6 & 1.6B\\
\textbf{XLM} & 387.8 $\pm$ 30.3 & 4.2 $\pm$ 1.7 & 550M\\
\textbf{XLNET\_base}  & 226.1 $\pm$  97.5 & 11.6 $\pm$ 7.9 & 110M \\
\textbf{XLNET\_large}  & 415.8 $\pm$ 53.2 & 4.3 $\pm$ 2.1 & 340M\\
\textbf{FAIR\_wmt19}  & 221.2 $\pm$ 66.6 & 14.6 $\pm$ 6.0 & 656M \\
\textbf{FAIR\_wmt20}  &  100.6 $\pm$  28.1 & 5.1 $\pm$ 3.0 & 749M \\
\textbf{TRANSFORMER\_XL}   & 211.7 $\pm$  53.9 & 9.8 $\pm$ 3.1 & 257M \\
\textbf{PPLM\_distil}  & 156.9 $\pm$ 40.1 & 10.7 $\pm$ 3.6 & 82M \\
\textbf{PPLM\_gpt2}   &  188.9 $\pm$  52.0 & 11.9 $\pm$ 4.5 & 124M \\
\bottomrule
\end{tabular}
}
\caption{Summary statistics of the \name~dataset.}
\label{tab:summary}
\end{table*}


    
    
    


\subsection{\name~Benchmark Tasks}

\paragraph{The Turing Test (TT) Task}
Our proposed Turing Test task aims to answer the question: \textit{Can we determine if a piece of text is human-written or machine-generated?}
This task is formulated as a binary classification problem with two labels -- \textit{human} and \textit{machine} -- modeled after the classical \textit{Turing Test} problem.
The \textit{Turing Test} examines the ability of a machine text-generator to exhibit intelligible behavior ascribed to humans. 
The goal is to build a model that causes the machine-generated texts to fail the \textit{Turing Test}.
Lastly, the TT task contains 19 subtasks with 19 human-machine pairs (e.g. GPT-2 XL vs. Human, GROVER\_base vs. Human, etc.).


\paragraph{The Authorship Attribution (AA) Task}
\textit{Authorship Attribution} is the identification and proper assignment of the author of a piece of text \cite{coyotl2006authorship}.
Our Authorship Attribution task aims to answer the question: 
\textit{If we determine that an article is human-written or machine-generated, 
can we further determine which neural language model generated all the articles that are said to be machine-generated?} 
This is a multi-class classification problem modeled after the traditional 
\textit{Authorship Attribution} problem.

\subsection{{\name} Benchmark Dataset}
We keep $168,612$ articles out of 200K after cleaning the text 
(see Appendix for data pre-processing \\details), and we build the benchmark dataset for each benchmark task - \textit{TT} and \textit{AA}. 
For the \textit{TT} task, there are 20 labels (i.e., 19 machine text-generators and 1 human), 
thus we can only have 19 pairs of human vs. machine. 
Therefore, we have 19 datasets for the TT task. 
To increase the difficulty of the \textit{TT} task, 
we cut each article in the test set in half, using only 50\% of the words. 
For the \textit{AA} task, we have 1 dataset containing all the labels. 
All datasets have train/validation/test sets which were split using the 70:10:20 ratio, respectively. 
To avoid topic bias, these sets were carefully split, such that all articles in the sets were unique to each other. Therefore, all articles generated by a prompt belonged only to one set. 

To make this dataset public, we added our datasets for each benchmark 
task and subtask to Hugging Face 
datasets\footnote{https://huggingface.co/datasets/\\turingbench/TuringBench/tree/main}. 
Figure~\ref{fig:code} demonstrates how to load the {\name} dataset.

\begin{figure}
    \centering

\begin{minipage}{0.45\textwidth}
\fontsize{5pt}{6pt}\selectfont
\begin{python}[fontsize=\small]
from datasets import load_dataset
import pandas as pd

# GPT-1 TT task
TT_gpt1 = load_dataset(
      'turingbench/TuringBench', 
      name='TT_gpt1', split='train')
TT_gpt1 = pd.DataFrame.from_dict(TT_gpt1)

# AA task
AA = load_dataset(
        'turingbench/TuringBench', 
        name='AA', split='train') 
AA = pd.DataFrame.from_dict(AA)
\end{python}
\end{minipage}
    \caption{Python code for loading the {\name} datasets using the Hugging Face API. 
    }
    \label{fig:code}
\end{figure}

\paragraph{Evaluation Metrics}
We use the traditional evaluation metrics such as: 
Precision, Recall, F1 score, and Accuracy to evaluate Machine/Deep Learning models for the benchmark tasks. 
However, for the TT tasks, we only use F1 scores since it is a more robust 
measure for the imbalanced datasets.

\begin{figure}[t!]
    \centering
    \includegraphics[width=0.5\textwidth]{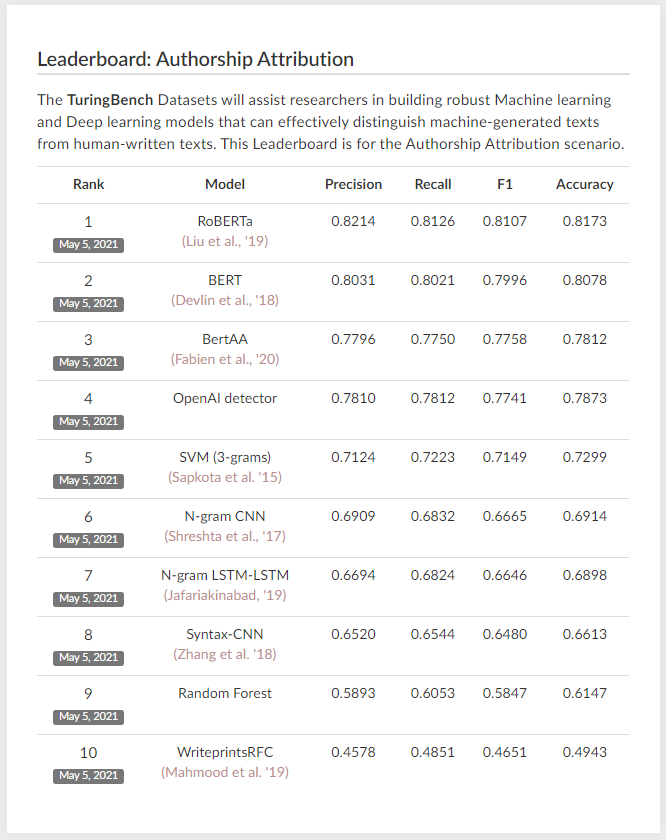}
    \caption{A screenshot of a leaderboard on the {\name} website.}
    \label{fig:web}
\end{figure}

\subsection{The Web Environment}
To create this {\name} environment, we built 2 versions of datasets - binary setting (i.e., \textit{human vs. GROVER-large, human vs. GPT-1, etc.}) for the TT tasks, 
and multi-class setting (i.e., \textit{human vs. GROVER-Large vs. GPT-1 vs. etc.}) for the AA task. 
To track progress, as shown in Figure~\ref{fig:web}, we create a website where each task and sub-task has its own
leaderboard that displays the evaluation metric
scores of models. 
Furthermore, to ensure the integrity of the process, even though contributors can obtain the {\name} datasets from Hugging Face 
datasets, we still ask contributors to 
submit their code and/or trained model weights for private testing. 
After testing, we update the website 
with the new models' scores.
Lastly, we rank the model performance using the F1 score from best to worst.

\begin{figure}[t!]
    \centering
    \includegraphics[width=0.5 \textwidth]{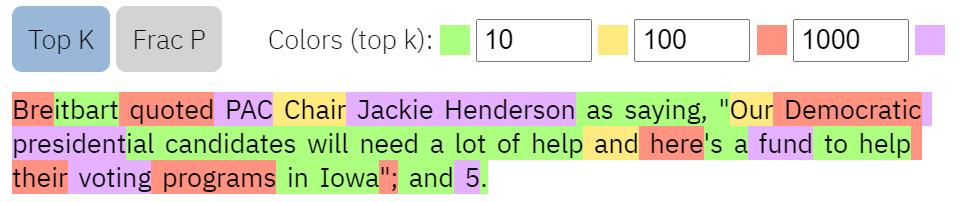}
    \caption{Using GLTR \cite{gehrmann-etal-2019-gltr} on a piece of text generated by GPT-3. 
    \textcolor{green}{Green} represents the most probable words; \textcolor{yellow}{yellow} the 2nd most probable;
    \textcolor{red}{Red} the least probable; and \textcolor{violet}{purple} the highest improbable words.
    Machine-generated texts are often populated with mostly \textcolor{green}{Green} and \textcolor{yellow}{yellow} words. However, we see that GPT-3-generated texts is very human-like.
    }
    \label{fig:gltr}
\end{figure}

\section{Experiments}

\begin{savenotes}
\begin{table*}[t!]
\centering
\footnotesize
\resizebox{\textwidth}{!}{
\begin{tabular}{p{0.2\linewidth} p{0.8\linewidth}}
\toprule
\textbf{TT Model} & \textbf{Description} \\
\midrule
GROVER detector & We use the GROVER-Large discriminator that is trained to detect GROVER-generated texts to predict the test labels. \\
GPT-2 detector & We use the trained 
weights
of RoBERTa-large  fine-tuned on GPT-2 XL outputs to predict the \textit{human} and \textit{machine} 
label of the test dataset. \\
GLTR  & In the GLTR demo, 
        the words are color coded to improve  human detection of
        machine-generated texts. Top 0-10 probable words are 
        \textcolor{green}{green}; 
        top 10-100 probable words are \textcolor{yellow}{yellow}; 
        top 100-1000 probable words 
        are \textcolor{red}{red} and top greater than 1000 words 
        are \textcolor{violet}{purple}. See Figure \ref{fig:gltr} 
        for an example of using GLTR
        and interpretation of its color schemes. 
        Thus, we define 
        human-written texts to be any article that 
        10\% or more of the words belong in the 
        top >1000 (i.e., \textcolor{violet}{purple} words).
\\
BERT & We fine-tune \textit{bert-base-cased}
         on the train set and classify on the test set.
         \\
RoBERTa  & 
        We fine-tune RoBERTa-base, a variant of BERT with the train set.
         \\
\bottomrule
\end{tabular}
}
\caption{Description of the Turing Test (TT) models.}
\label{tab:tt_baseline}
\end{table*}
\end{savenotes}

\begin{savenotes}
\begin{table*}[t!]
\centering
\footnotesize
\resizebox{\textwidth}{!}{
\begin{tabular}{p{0.26\linewidth} p{0.74\linewidth}}
\toprule
\textbf{AA Model} & \textbf{Description} \\
\midrule
Random Forest & Using TF-IDF to represent the data, we classify the texts with Random Forest. \\
SVM (3-grams) & We represent the texts as 3-grams and classify the texts with SVM. \\
WriteprintsRFC & Writeprints features + Random Forest Classifier. \\
OpenAI detector & We re-purposed RoBERTa-base (\textit{roberta-base-openai-detector}) model that was originally fine-tuned on GPT-2 XL outputs to detect machine-generated texts, by training the model as a multi-classifier for the AA task. \\
Syntax-CNN & Use Part-Of-Speech to capture the syntax of the texts and classify the texts with CNN \\
N-gram CNN & Represent the data with n-grams (uni-grams) and classify texts with CNN \\
N-gram LSTM-LSTM & Represent the data with n-grams (uni-grams) and classify texts with LSTM \\
BertAA & Using BERT + Style + Hybrid features to achieve automatic authorship attribution. Style features include:
\textit{length of text, 
number of words, average length of words,} etc. and Hybrid features include: \textit{frequency of the 100 most frequent
character-level bi-grams} and \textit{the 100 most frequent
character-level tri-grams}. \\
BERT-Multinomial & Using BERT for multi-class classification \\
RoBERTa-Multinomial & Using RoBERTa for multi-class classification\\
\bottomrule
\end{tabular}
}
\caption{Description of the Authorship Attribution (AA) models.}
\label{tab:aa_baseline}
\end{table*}
\end{savenotes}

We experiment with several SOTA and baseline models as summarized in Table~\ref{tab:tt_baseline} for \textbf{Turing Test} and Table~\ref{tab:aa_baseline} for \textbf{Authorship Attribution}, and Table~\ref{tab:tt} and Table~\ref{tab:aa} show their results. 


\begin{table*}[t!]
\centering
\footnotesize
\resizebox{15cm}{!}{
\begin{tabular}{l|cc|ccccc|c}
\toprule
\textbf{Human vs. }  & \textbf{Human Test} & \textbf{Human Test} &\textbf{GROVER} & \textbf{GPT-2} & \textbf{GLTR} & 
\textbf{BERT} & \textbf{RoBERTa} & AVG\\

  & \textbf{(machine)} & \textbf{(human vs. machine)} & {\bf detector}& {\bf detector} &  & 
 &  & \\
\midrule 
\textbf{GPT-1} & 0.4000 & 0.5600 & 0.5792 & 0.9854 & 0.4743 &  0.9503 &  0.9783 & 
0.7935\\
\textbf{GPT-2\_small} & 0.6200 & 0.4400 & 0.5685 & 0.5595 & 0.5083 &  0.7517 &   0.7104 & 0.6197 \\
\textbf{GPT-2\_medium} & 0.5800 & 0.4800 & 0.5562  & 0.4652 & 0.4879 &  0.6491  &  0.7542  & 0.5825 \\
\textbf{GPT-2\_large} & 0.7400 & 0.4400 & 0.5497 & 0.4507&  0.4582 &  0.7291 & 0.7944  & 0.5964 \\
\textbf{GPT-2\_xl} & 0.6000 & 0.4800 & 0.5549  &  0.4209 & 0.4501 &  0.7854   & 0.7842  & 
0.5991 \\
\textbf{GPT-2\_PyTorch} & 0.5000 & 0.5600 & 0.5679 & 0.5096 & 0.7183 &  0.9875 & 0.8444 &  0.7255 \\
\textbf{GPT-3} & 0.4400 & 0.5800 & 0.5746  & 0.5293 & 0.3476 &  0.7944 &  0.5209  & 
\underline{0.5534}\\
\textbf{GROVER\_base} & 0.3200 & 0.4200 & 0.5766 &  0.8400 &  0.3854 & 0.9831 & 0.9870  & 
0.7544 \\
\textbf{GROVER\_large} & 0.4800 & 0.5800 & 0.5442  &  0.5974 &  0.4090   & 0.9837  & 0.9875  & 0.7044   \\
\textbf{GROVER\_mega} & 0.5400 & 0.4800 & 0.5138 & 0.4190 &   0.4203 & 0.9677 &  0.9416   &  0.6525 \\
\textbf{CTRL} & 0.5000 & 0.6900 & 0.4865 &  0.3830 & 0.8798  &   0.9960 &  0.9950 &  
0.7481 \\
\textbf{XLM} & 0.6600 & 0.7000 & 0.5037  &  0.5100 &  0.8907 &  0.9997  & 0.5848   &   
0.6978 \\
\textbf{XLNET\_base} & 0.5200 & 0.5400 & 0.5813  & 0.7549 & 0.7541 &  0.9935 &   0.7941  &  
0.7756 \\
\textbf{XLNET\_large} & 0.5200 & 0.5200  & 0.5778  & 0.8952 &  0.8763  &  0.9997 &   0.9959 &   0.8690 \\
\textbf{FAIR\_wmt19} & 0.5600 & 0.5600 & 0.5569  & 0.4616 & 0.5628 & 0.9329   & 0.8434  &  
0.6715 \\
\textbf{FAIR\_wmt20} & 0.5800 & 0.2800 & 0.5790 & 0.4775 & 0.4907 & 0.4701  & 0.4531  & \textbf{0.4941} \\
\textbf{TRANSFORMER\_XL} & 0.5000 & 0.5000 & 0.5830  & 0.9234 & 0.3524   &  0.9721 &  0.9640  & 0.7590 \\
\textbf{PPLM\_distil} & 0.5600 & 0.4400 & 0.5878  & 0.7178 &  0.6425 & 0.8828  &  0.8978 &  0.7457 \\
\textbf{PPLM\_gpt2} & 0.5600 & 0.5000 & 0.5815  & 0.5602 & 0.6842 &  0.8890 &0.9015  &  0.7233 \\
\midrule
AVG & 0.5358 & 0.5132 & 0.5591 & 0.6032 & 0.5681  & \textbf{0.8799} & \underline{0.8280} &  \\
\bottomrule
\end{tabular}
}
\caption{Compared Human Test vs. 
Test F1 scores of Turing Test models (\textbf{bold} and \underline{underlined} are \#1 and \#2 performance, respectively).
Human Test (machine) asked humans to decide if a given article is machine-generated or not, while Human Test (human vs. machine) asked humans which of the two given texts is machine-generated.}
\label{tab:tt}
\end{table*}

\begin{table}[t!]
\centering
\footnotesize
\resizebox{8cm}{!}{
\begin{tabular}{lcccc}
\toprule
\textbf{AA Model} & \textbf{P} & \textbf{R} & \textbf{F1} & \textbf{Accuracy} \\
\midrule
Random Forest &  0.5893  & 0.6053 &  0.5847 & 0.6147 \\
SVM (3-grams) & 0.7124  &  0.7223  &  0.7149  & 0.7299   \\
WriteprintsRFC & 0.4578 & 0.4851  & 0.4651 & 0.4943 \\
OpenAI detector &  0.7810  & 0.7812  & 0.7741 & 0.7873 \\
Syntax-CNN & 0.6520 & 0.6544 & 0.6480 & 0.6613 \\
N-gram CNN & 0.6909 & 0.6832 & 0.6665 & 0.6914 \\
N-gram LSTM-LSTM & 0.6694 &  0.6824 & 0.6646 & 0.6898 \\
BertAA & 0.7796 & 0.7750 & 0.7758 & 0.7812 \\
BERT-Multinomial &  \underline{0.8031}  & \underline{0.8021}  &  \underline{0.7996} &  \underline{0.8078}  \\
RoBERTa-Multinomial &  \textbf{0.8214}  &  \textbf{0.8126}  &   \textbf{0.8107}   & \textbf{0.8173} \\
\toprule
\end{tabular}
}
\caption{Performance of Authorship Attribution models (\textbf{bold} and \underline{underlined} are \#1 and \#2 performance, respectively).}
\label{tab:aa}
\end{table}

\subsection{Results from Turing Test}

The \textit{Turing Test} task is formulated as a binary classification problem with \textit{human} and \textit{machine} labels.
In order to make the TT task even more difficult, we train and validate on the full articles 
generated by the text-generators and test on only 50\% of the words of each article in the test set.
We intend to capture the differences that will exist between train and 
test data in the real world in this scenario.

We compare 3 SOTA TT models - GROVER detector~\cite{zellers2019defending}, GPT-2 detector~\cite{solaiman2019release}, and GLTR~\cite{gehrmann-etal-2019-gltr}.
We observe in Table~\ref{tab:tt} that the average F1 scores are 0.56,
0.60, and 0.57, respectively. 
Next, using other text classifiers such as BERT~\cite{devlin2018bert} and RoBERTa~\cite{liu2019roberta} brings a significant improvement in F1 scores (0.85 for both BERT and RoBERTa).

This performance improvement occurs mainly because BERT and RoBERTa are fine-tuned with the train set
of each TT subtasks, while the TT models' pre-trained models were used to classify the test set 
without any further training. 

Additionally, averaging over all the 5 TT models, we find that FAIR\_wmt20 and GPT-3, 
the most recent text-generators in the list,
achieve the lowest average F1 score (0.49 and 0.55), thus making them the language models that produce the most indistinguishable texts, while XLNET\_large
has the highest average F1 score (0.87) using all TT models. 
XLNET has a high F1 score because it 
implements a text padding technique for generation which often 
negatively affects the generation quality. 

We also run two human experiments using the Amazon Mechanical Turk (AMT) environment, recruiting workers with at least 95\% approval rate of Human Intelligence Task (HIT).
In the experiments,
we randomly sampled 50 articles per each language model (across all 19 models) and performed two tests, where workers (1) vote if a given article is machine-generated or not, and (2) vote which of two given articles is machine-generated. 
 These experiments yielded the AVG-accuracies of 0.535 and 0.513  (random-guess=0.5), respectively.

This part of experiments was reviewed and approved by the Institutional Review Board of our institution.

\subsection{Results from Authorship Attribution}
Since there are 20 labels in AA, the chance performance is at 0.05 (i.e., 5\% in accuracy).
Due to this difficulty, we use the full article contents in the test set. 
We compare different SOTA and popular techniques for automatic authorship attribution for 
our AA task including 
Random Forest, 
SVM (3-grams)~\cite{sapkota2015not}, 
WriteprintsRFC~\cite{mahmood2019girl},
OpenAI detector\footnote{https://huggingface.co/roberta-base-openai-detector}, 
Syntax-CNN~\cite{zhang2018syntax}, N-gram CNN~\cite{shrestha2017convolutional}, 
N-gram LSTM-LSTM~\cite{jafariakinabad2019syntactic},
BertAA~\cite{fabien2020bertaa},
BERT-Multinomial~\cite{devlin2018bert}, 
RoBERTa-Multinomial~\cite{liu2019roberta}.
We find that BERT and RoBERTa outperform all the 
AA models, sometimes significantly, achieving the F1  scores of 0.80 and 0.81, respectively. 

Interestingly, we observe that OpenAI detector, a RoBERTa-base model fine-tuned on GPT-2 XL outputs, 
does not outperform BERT-Multinomial and RoBERTa-Multinomial for 
this AA task although it 
performs comparatively, achieving a 0.77 as F1 score. BertAA achieves a slightly better F1 score (0.78).


\section{Discussion}
We present several observations from our experimental results.

\begin{figure}[t!]
    \centering
    \hspace{-15pt}
    \includegraphics[width=1.04\linewidth]{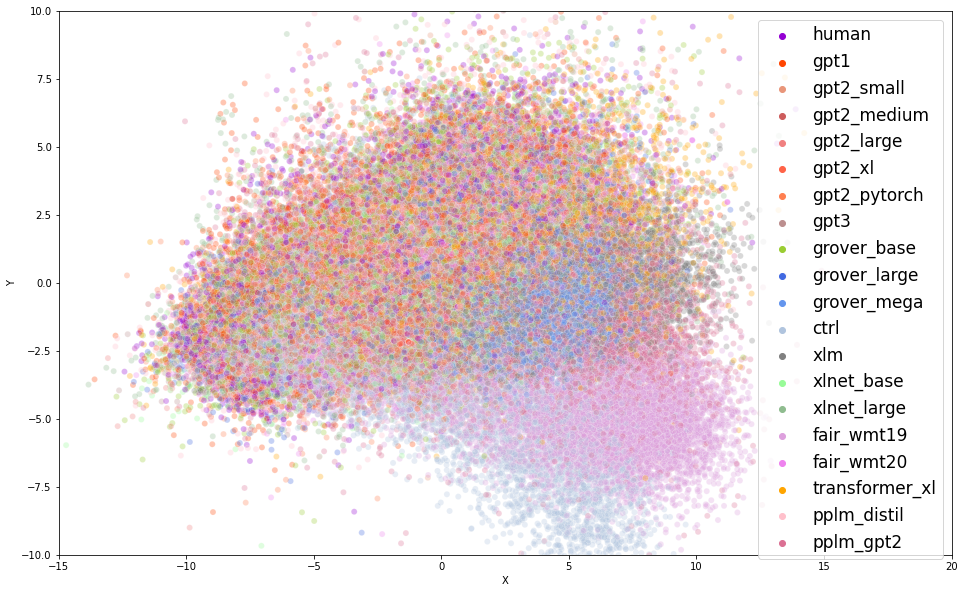}
    \caption{PCA plot of psycho-linguistics features of the 
    {\name} dataset, using LIWC to attempt to capture the stylistic 
    signatures of the dataset
    }
    \label{fig:pca}
\end{figure}

\begin{figure}[t!]
    \centering
    \includegraphics[width=0.95\linewidth]{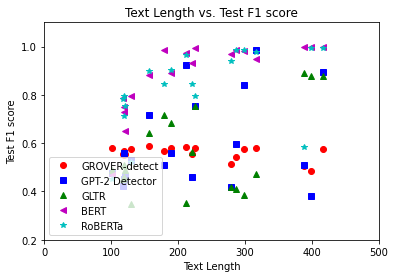}
    \caption
    {Despite the varying lengths 
    of the generated texts (100--400) in Table \ref{tab:summary}, no   correlation between text length and F1 score was found.}
    \label{fig:lengthF1}
\end{figure}

\noindent
\begin{enumerate}[leftmargin=\dimexpr\parindent+0.1\labelwidth\relax]
    \item \textbf{Both TT and AA tasks are non-trivial:} 
    The average F1 score for each human vs. machine subtask and TT model is 
    below 0.87, with FAIR\_wmt20 achieving the lowest (0.49).
    FAIR\_wmt20 is the newest text-generator in our list and before that
    we have GPT-3 which achieves the
    second lowest average F1 score (0.55).
    This suggests a trend that as newer text-generators get built, generated texts will 
    become even more human-like, making the TT and AA tasks more difficult. 
    
    Additionally, the difficulty of the AA task is further demonstrated by the PCA plot of  
    linguistic features LIWC of the {\name} dataset in Figure~\ref{fig:pca}.
    Using LIWC to capture stylistic signatures 
    of authors has been studied~\cite{goldstein2009person,uchendu-etal-2020-authorship}. 
    However, we observe that there are quite a few overlaps in linguistic features across different authors (i.e., language models).
    This makes these authors' writing styles linearly inseparable. 
    
    \item \textbf{No one size fits all:} 
    We observe in Table~\ref{tab:tt} that there is no one detection model that performs well across all 20 TT tasks. 
    For instance, while BERT achieved the highest average F1 score, 
    it still underperformed in detecting FAIR\_wmt20.   
    However, GROVER detector achieved a highest F1 score in detecting FAIR\_wmt20. 
    
    \item \textbf{Humans detect machine-generated texts at chance level:}
First two columns of Table 5 show the results of human detection test.
    In the first AMT-based tests, we randomly sampled 50  machine-generated texts and asked humans to decide if the given text is human-written  or machine-generated (i.e., humans do not know whether they are shown only machine-generated texts in the test). In the second test, we showed two texts at ramdom, one written by humans and the other generated by machines, and
    asked humans to decide which of the two are machine-generated (i.e., humans know that at least one of two is machine-generated). 
    
    Based on the average accuracies of two human tests, by and large, we observe that humans  currently differentiate machine-generated 
    texts from human-written ones, not much better (i.e., 0.535 and 0.513) than the level of random guessing (i.e., 0.5).
    
    \item \textbf{Not all text-generators are created equal:} 
    As shown in Table~\ref{tab:tt}, the average F1 score for each human vs. machine subtask and TT model is below 0.87, with FAIR\_wmt20 achieving the lowest (0.49).
    Consequently, this suggests that FAIR\_wmt20 is the most sophisticated text-generator 
    and thus, the hardest to detect. 
    Other generators that are also hard to detect based on their < 0.62 F1 scores are: 
    GPT-3, GPT-2\_small, GPT-2\_medium,
    GPT-2\_large, and GPT-2\_XL.
    
    \item \textbf{Sophisticated machine-generated texts often get detected as human-written:} 
    We observe an interesting phenomenon with these SOTA 
    TT models. For instance, even though the labels in the binary 
    classification task are approximately evenly split,
    GPT-2 detector and 
    GLTR achieve below F1 score of 0.4 in some subtasks. 
    This happens because TT models do not generalize well 
    to those specific text-generators (i.e., GROVER\_base, CTRL, GPT-3,
    TRANSFORMER\_XL) and mistakenly predicts the majority of the texts as
    \textit{human-written}.

    \item \textbf{TT models do not always perform as expected:} 
    While both GROVER  and
    GPT-2 detectors are trained to detect GROVER-generated and 
    GPT-2-generated texts, respectively, they underperform
    in detecting those texts. For instance, GROVER detector
    performs the best in detecting PPLM\_distil and 
    PPLM\_gpt2 texts, while GPT-2 detector
    performs significantly better at detecting GPT-1, TRANSFORMER\_XL and 
    XLNET\_large texts.  

    \item \textbf{Length of texts does not affect model performance:} 
    Due to the varying length of texts (i.e. 100-400)
    in Table~\ref{tab:summary}, 
    we plot the length of generated texts 
    vs. the F1 scores of TT models in Figure~\ref{fig:lengthF1}.
    However, the figure suggests that there is no clear correlation
    between model performance and length of texts for all models except RoBERTa. This suggests that RoBERTa performance is text length-dependent. 
    
    \item \textbf{Traditional AA models cannot fully capture an author's style ``yet'':} 
    SOTA  AA models cannot capture all of the stylistic features of 
    human and machine text-generators. From Figure~\ref{fig:pca} we observe that the
    psycho-linguistic features of the 20 authors in the {\name} dataset are too 
    similar, causing  them to overlap in the plot. This suggests that 
    machine-generated texts are becoming more similar to human-written texts in styles.
    
    Therefore, traditional ways to capture an author's writing style will no 
    longer be sufficient to achieve accurate automatic authorship attribution. 
    This is further confirmed in the performance of classical AA models such as SVM and Random Forest. 
    Similarly, we find that even deep learning based AA models are still unable to fully
    capture the distinct writing styles of all 20 authors. 
    These results suggest that one needs to develop a model that can unearth 
   more subtle yet distinct patterns that exist across 20 models.

    \item \textbf{Humans have a wide writing style range:}
      In Figure \ref{fig:pca}, we observe that 
      human-written features spread out all over the plot, while all  machine-generated texts stay in little pockets of the plots. This suggests that humans may have a wider range of writing levels/styles, while  machines have a more limited range of writing levels/styles (e.g., high school to college).

\end{enumerate}

\section{Future Work}
The experimental results suggest that we need better models to solve both TT and AA tasks as traditional AA models/features alone cannot solve the AA or TT problem. In addition, black-box detectors may be no longer sufficient for detection, as it cannot explain ``why" a text is machine-generated or human-written yet. 
A research direction that GLTR-like framework points at may be able to help capture the nuanced nature of neural text-generators better.
In addition, in future, a more complicated situation may emerge where a user may  
generate different parts of an article using different neural text-generators to intentionally mask the writing style of the generated text, thus confusing  detectors--i.e., \textit{Authorship Obfuscation}.

\section{Conclusion}
In this paper, we have introduced the {\name} environment and its preliminary results for both Turing Test (TT) and Authorship Attribution (AA) tasks. While varied, overall, (1) many contemporary language models can generate texts whose qualities are,  to human eyes, indistinguishable from human-written texts, and (2) while some computational solutions for both TT and AA tasks can differentiate human-written texts from machine-generated ones much better than random guessing, overall, the community needs to research and develop better solutions for mission-critical applications.
We hope that the {\name} environment will provide a platform on which insights 
into ways to tackle this urgent issue can be developed and shared. 

\section{Ethics Statement}
We build {\name} by collecting public human-written news
articles (mostly politics), and use the \textit{Titles} of these articles as \textit{Prompts} to 
generate similar news articles with neural text-generators. 
Some of these human-written articles were scraped from CNN and Washington Post news websites, and others from Kaggle. See Appendix for links to Kaggle datasets. 
However, 
while the purpose of the {\name} environment is to call attention to the urgent need for 
detectors of machine-generated texts, the potential negative uses of this research are not lost on us.

We understand that the insights we provide in this work can be used maliciously to thwart 
the performance of these detectors. Also, since we have released our dataset publicly, 
we understand that malicious users can copy the political articles generated by neural text-generators such as GPT-3, make minor changes, and post them online under the guise of real news. However, we 
believe that this work will lead to the creation of strong detectors of machine-generated texts, so 
that even human-edited machine-generated texts will still be detected in future.

\section*{Acknowledgments}
This work was in part supported by NSF awards \#1742702, \#1820609, \#1909702, \#1915801, and \#2114824.

\bibliographystyle{acl_natbib}
\bibliography{anthology,custom}

\clearpage
\appendix
\section{Appendices} \label{sec:appendix}

\subsection{Data Generation Implementation}
Generating texts with these Language models is very computationally 
expensive. Some of the python code used to generate the texts were not 
written for large scale generation, so we had to re-purpose it for 
our task. We mostly used Google Colab pro's GPU - 12GB NVIDIA Tesla K80
to generate our texts. However, since \textit{PPLM} was the heaviest 
language model computationally, we used a machine with more GPUs - 
NVIDIA Tesla K80s and P100s. 

Most generators took 24 -- 72 hours to generate 10K articles. However, 
\textit{PPLM} took about 430 hours for \textit{PPLM\_distil} and 
about 600 hours for \textit{PPLM\_gpt2}. It is important to note that 
probably a few coding choices could reduce the computational cost of 
running \textit{PPLM}, we just did not get to it. 
See the description of building the human dataset and 10 language model architectures used to generate the rest of the dataset.
The table also contains the links to the dataset and github repo of 
some of the models. 

\subsection{Data Pre-processing}
Some of the generated texts contain non-English tokens such as 
$\langle UNK\rangle$, $\langle eos\rangle$, $\langle eod\rangle$, $\langle eop\rangle$, $\langle |endoftext|\rangle$, etc. which we removed. 
Also, in an attempt to generate texts with the specified word count (i.e., 400), 
some of the generators had a tendency to repeat a particular word multiple 
times consecutively. This introduced bias 
into our Machine Learning models, making it easier to detect 
such generated texts. Therefore, we removed words that were repeated 
consecutively, leaving only one. Next, those same text-generators
also had a tendency to generate texts where a random word would 
have the last character repeated multiple times. For instance, 
a word like ``expressed", could be spelt like 
``expresseddddddddddddddddddddddddddddd''. 
This also made such generators easy to detect, 
so we removed words more than 20 characters to get 
rid of such words. Lastly, the word ``CNN'' was used 
heavily by a few generators, making it easier 
to detect such generators. Therefore, 
we removed the word, ``CNN" from all the articles.


\begin{table}[t!]
\centering
\footnotesize
\resizebox{7cm}{!}{
\begin{tabular}{lc}
\toprule
\textbf{Text Generator}  & \textbf{\# of Data samples}\\
\midrule
\textbf{Human}  &   8,854\\
\textbf{GPT-1} &  8,309 \\
\textbf{GPT-2\_small} & 8,164  \\
\textbf{GPT-2\_medium} & 8,164 \\
\textbf{GPT-2\_large} & 8,164 \\
\textbf{GPT-2\_xl} &  8,309 \\
\textbf{GPT-2\_PyTorch} & 8,854\\
\textbf{GPT-3} & 8,164 \\
\textbf{GROVER\_base} & 8,854\\
\textbf{GROVER\_large}  & 8,164 \\
\textbf{GROVER\_mega}  & 8,164\\
\textbf{CTRL} & 8,121 \\
\textbf{XLM} & 8,852 \\
\textbf{XLNET\_base}  &  8,854\\
\textbf{XLNET\_large}  & 8,134\\
\textbf{FAIR\_wmt19}  & 8,164 \\
\textbf{FAIR\_wmt20}  &  8,309 \\
\textbf{TRANSFORMER\_XL} & 8,306 \\
\textbf{PPLM\_distil}  &  8,854\\
\textbf{PPLM\_gpt2}   &   8,854\\
\bottomrule
\end{tabular}
}
\caption{\# of data samples in the \name~dataset}
\label{tab:data_distribution}
\end{table}

Before pre-processing of the data, we had 200K, and after the 
process, we have $168,612$. See data distribution in 
Table \ref{tab:data_distribution} of the cleaned dataset. 
We can observe that the distribution of the dataset is 
still approximately the same. 

\begin{savenotes}
\begin{table*}[th!]
\footnotesize
\resizebox{\textwidth}{!}{
\begin{tabular}{p{0.25\linewidth} p{0.75\linewidth}}
\toprule
\textbf{TEXT-GENERATORS} & \multicolumn{1}{c}{\textbf{DESCRIPTION}} \\
\midrule
Human  & We collected news titles (mostly Politics) and contents from CNN, Washington Post, and Kaggle  
\footnote{https://www.kaggle.com/snapcrack/all-the-news,} \footnote{https://www.kaggle.com/sunnysai12345/news-summary} \footnote{https://www.kaggle.com/ryanxjhan/cbc-news-coronavirus-articles-march-26} \footnote{https://www.kaggle.com/patjob/articlescrape}
. 
Next, we removed articles that did not have the desired word length (i.e., 200--500). This resulted in 130K articles, but only 10K was used for the article generations.
\\ \midrule
GPT-1    &  
Texts are generated with the huggingface github repo\footnote{https://github.com/huggingface/transformers}.\\
GPT-2   & We use 4 GPT-2 pre-trained models - PyTorch model \footnote{https://github.com/graykode/gpt-2-Pytorch}, small (124 million parameters), medium (355 million parameters), large (774  million parameters), and extra-large (1558 million  parameters) \footnote{https://github.com/minimaxir/aitextgen} to generate texts.  \\
GPT-3 & Texts are generated with the OpenAI GPT-3 API using the \textit{davinci} engine. \\
GROVER  & We use code from repo\footnote{https://github.com/rowanz/grover} to generate from Grover's 3 pre-trained models: \textbf{GROVER-base}, \textbf{GROVER-large}, \textbf{GROVER-mega}. \\
CTRL & Conditional Transformer Language Model For Controllable Generation \footnote{https://github.com/salesforce/ctrl} uses control codes to guide generation. We use \textit{News} control code to generate long articles.\\
XLM & We generated texts using huggingface repo. \\
XLNET  & We generated texts with: 2 XLNET pre-trained models: \textbf{XLNET-base}, and \textbf{XLNET-large} using huggingface.\\
FAIR\_wmt & We use two Facebook's FAIR English models - \textbf{wmt19}\footnote{https://github.com/pytorch/fairseq/tree\\/master/examples/wmt19}  and  \textbf{wmt20}\footnote{https://github.com/pytorch/fairseq/tree\\/master/examples/wmt20} to generate texts with FAIRSEQ sequence modeling toolkit.\\
TRANSFORMER\_XL & We generated texts with this language model's setup on huggingface.\\
PPLM & PPLM fuses GPT-2's pre-trained model with bag of words to generate more specific texts. We used the \textit{Politics} bag of words model to generate texts', using the code\footnote{https://github.com/uber-research/PPLM}, and used the perturbed version. Next, we fused PPLM with two pre-trained models (i.e., distilGPT-2, and GPT-2) and generated texts with them, forming: \textbf{PPLM\_distil}, \textbf{PPLM\_gpt2}. These models are gotten from the huggingface model repository\footnote{https://huggingface.co/models}. \\
\bottomrule
\end{tabular}
}
\caption{Description of the Text-generators in the \name~dataset.}
\label{tab:desc2}
\end{table*}
\end{savenotes}


\begin{figure*}
    \centering
    \includegraphics[width=0.9\linewidth]{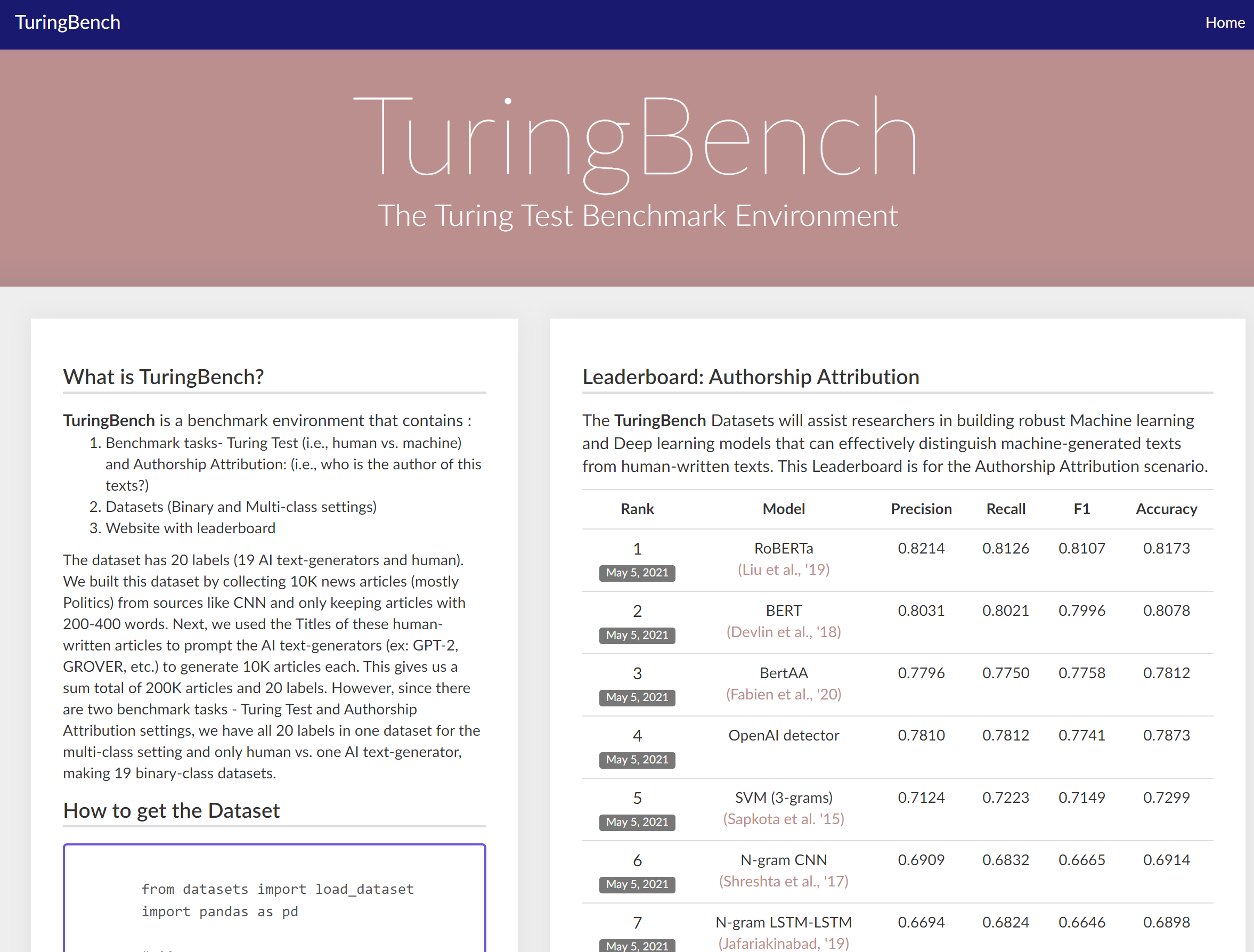}
    \caption{{\name} website interface}
    \label{fig:inter}
\end{figure*}

\subsection{\name~Website}
We create the {\name} website using the SQuAD website framework. 
The website contains a description of the benchmark datasets and 
benchmark tasks. Each benchmark task has a leaderboard that 
shows the models used to solve the tasks. These models are 
rated from best to worst. For the AA tasks, we use the standard 
Machine learning evaluation metrics such as: 
\textit{Precision}, \textit{Recall}, \textit{F1 score}, and \textit{Accuracy}. And we use only \textit{F1 score} for the 
TT task because it is a binary classification problem and \textit{F1 score}
is sufficient for the problem. See website interface in Figure \ref{fig:inter}.

\begin{table}[t!]
    \centering
    \begin{tabular}{c|c}
    \hline
        \textbf{Model} & \textbf{Run-time}  \\
        \hline
       GROVER detector  &  25 -- 30 minutes \\
       GPT-2 detector & 5 -- 10 minutes \\
       GLTR & 4 -- 5 hours \\
       BERT & 25 -- 40 minutes \\
       RoBERTa & 45 -- 1 hour \\
       \hline
    \end{tabular}
    \caption{TT model Run-time per task}
    \label{tab:tt_run}
\end{table}

\subsection{Experiments}
All experiments, except \textit{GLTR} and \textit{GPT-2 detector} were done using the Google colab pro's GPU stated above.
Experiments with \textit{GLTR} and \textit{GPT-2 detector} were done 
in a machine with 4 GPUs - NVIDIA Quadro RTX 8000. 

\subsubsection{TT models}
Each of the models used their default hyperparameters. There was no 
hyperparameter tuning performed. 
We used 
GROVER-Large discriminator for \textit{GROVER detector}, the weights of 
roberta-large fine-tuned on GPT-2 XL outputs for \textit{GPT-2 detector},
and GPT-2 117M model for \textit{GLTR}. None of these models were trained on our dataset. We tested their performance on predicting on our test set.
Next, we fine-tuned BERT and RoBERTa on our train set and validate these 
models on our validation set for each TT task. BERT was fine-tuned for 3
epochs and RoBERTa, 3--5 epochs with $2e^{-5}$ learning rate. See Table 
\ref{tab:tt_run} for run-time of the models.

\subsubsection{AA models}
We used the default hyperparamters of the AA models for the AA task. 
Also, we did not perform any hyperparameter tuning on these models. 
\textit{Random Forest} and \textit{SVM} take about 
30 minutes -- 1 hour to converge. 
\textit{WriteprintsRFC} took about 15 minutes to converge. 
\textit{Syntax-CNN}, \textit{N-gram CNN}, and \textit{N-gram LSTM-LSTM}
took about 30 minutes to converge. \textit{OpenAI detector}
took about an hour to converge. \textit{BERT-Multinomial} and
\textit{RoBERTa-Multinomial} took about 1 -- 2 hours to converge. 
\textit{BertAA} took about 5 hours to converge.

\end{document}